\def\year{2021}\relax
\documentclass[letterpaper]{article} % DO NOT CHANGE THIS
\usepackage{aaai21}  % DO NOT CHANGE THIS
\usepackage{times}  % DO NOT CHANGE THIS
\usepackage{helvet} % DO NOT CHANGE THIS
\usepackage{courier}  % DO NOT CHANGE THIS
\usepackage[hyphens]{url}  % DO NOT CHANGE THIS
\usepackage{graphicx} % DO NOT CHANGE THIS
\usepackage{natbib}  % DO NOT CHANGE THIS AND DO NOT ADD ANY OPTIONS TO IT
\usepackage{caption} % DO NOT CHANGE THIS AND DO NOT ADD ANY OPTIONS TO IT
\usepackage{amssymb}
\usepackage{algorithm2e}
\usepackage{algorithmicx}

\usepackage{changes}

\title{A MultiModal Social Robot Toward Personalized Emotion Interaction}
\author {
    Baijun Xie and
    Chung Hyuk Park \\
}
\affiliations {
     Department of Biomedical Engineering \\
     School of Engineering and Applied Science \\ 
     The George Washington University, U.S.A.\\
    bdxie@gwu.edu, chpark@gwu.edu
}
\begin{document}

\maketitle

\begin{abstract}
Human emotions are expressed through multiple modalities, including verbal and non-verbal information. Moreover, the affective states of human users can be the indicator for the level of engagement and successful interaction, suitable for the robot to use as a rewarding factor to optimize robotic behaviors through interaction. This study demonstrates a multimodal human-robot interaction (HRI) framework with reinforcement learning to enhance the robotic interaction policy and personalize emotional interaction for a human user. The goal is to apply this framework in social scenarios that can let the robots generate a more natural and engaging HRI framework.
\end{abstract}

\section{Introduction}
This paper presents an ongoing study on multimodal human-robot interaction (HRI) with a reinforcement learning (RL) dialogue agent. To develop an effective HRI system for social robots that can naturally interact with human users, the robots need to accurately identify the user's affective states and respond to the users with personalized behaviors accordingly. Moreover, robots can also improve the user's engagement in the long-term interaction~\cite{leite2013social} through personalized interaction policy. However, in reality, emotion recognition could be challenged in HRI since humans convey information and feelings via different sources of social cues such as facial expression, body language, and speech. Thus, the studies of multimodal emotion recognition have attracted considerable interest in recent years. 

These fusion strategies in the previous years can be summarized into three main categories: feature-level fusion, decision-level fusion, and model-level fusion~\cite{wu2014survey}. \citet{schuller2012avec} demonstrated a baseline model in the Audio/Video Emotion Challenge (AVEC) 2021, which concatenated the audio and visual features using feature-level fusion and then used support vector regression to predict the continuous affective values. On the other hand, the decision-level fusion can process different types of inputs with diverse classifiers, and the final estimation can fuse the outputs from all classifiers. For example, the posterior probabilities from the predictions of the audio and video classifiers can be combined to get the final estimations~\cite{schuller2011avec}. Deep learning models, such as Long-Short Term Memory Recurrent Neural Network (LSTM-RNN), have been used to achieve a model-level fusion~\cite{chen2016multi}. More recently, researchers have leveraged the transformer to fuse different modalities of inputs~\cite{tsai2019multimodal, xie2021robust}, where the transformer is proposed by~\citet{vaswani2017attention} for solving Sequence-to-Sequence (Seq2Seq) problem based on attention mechanism only without any recurrent structure as RNN.

Once the robots have the ability to recognize the user's affective states, the HRI system can utilize this information as inputs and determine the behaviors of the robots. The utilization of emotional models in HRI can create more natural and engaging HRI experiences, as evidenced by Ficocelli et al.~\cite{ficocelli2015promoting}. The developed emotional model can also be used for the empathetic appraisal for social robots that can interact with children in the long-term study~\cite{leite2014empathic}. One recent study also presents a social robot with emotional interaction that can elicit particular emotions using non-verbal emotional behaviors~\cite{shao2020user}. Thus, we believe that the affective states of the human user can be an effective indicator for the HRI system to generate more engaging and natural behaviors for the interaction.

Moreover, in order to develop a more natural HRI framework, the robots need to learn human preferences and skills. Recent studies utilize RL to let the robots learn social skills through the interactions~\cite{kim2017intrinsic, qureshi2018intrinsically}. Furthermore, \citet{ritschel2018socially} proposed an RL framework that can adapt robot behavior using social signals. Robots' capacity to learn multimodal cues adaptively and associate them with the context is recognized as being the vital factor in HRI. \citet{cui2020empathic} proposed a novel data-driven framework, EMPATHIC, which aims to improve the policy of the agent in learning tasks by using implicit human facial reactions feedback. However, there is still a need for the study on the RL agent for multimodal HRI. Thus, in this study, we present a multimodal HRI system with an RL framework that can accept multiple modalities of inputs to shape the reward signal and adapt the robot behaviors to generate more positive feedback for personalized interaction with the user. For the rest of the paper, We first present the developed dialogue RL framework in our system cause the speech will be the primary social cue for our robot platform. Then, we depict the integral diagram of our multimodal HRI system. Finally, we propose the evaluation plan for our future study.

\section{Related Studies}
Recent studies in multimodal HRI are providing enhanced abilities for emotional interaction~\cite{liu2016multimodal, hong2020multimodal, li2019expressing}, where the emotional abilities, including understanding users' affective states and expressing emotional behaviors accordingly. In the study proposed by~\citet{liu2016multimodal}, multiple modalities, including vocal, facial, and gesture features, were combined to detect the emotions by using a Bayesian classifier. The Nao humanoid robot was used as the platform to convey emotional behaviors by mimicking the user's behaviors. \citet{hong2020multimodal} proposed a multimodal emotional HRI architecture that can interpret human emotional states via multimodal social cues and promote natural bidirectional communications with users. Gestures and vocal features were fused to detect the user's affect and the robot utilized the user's affect to express the corresponding emotional behaviors. Another study by~\citet{li2019expressing} also used emotions for a spoken dialogue system in multimodal HRI which aimed to conduct the natural conversation. The study proposed an emotion recognition model which combines valence from prosody and sentiment analysis for the robot to express reactive emotions adequately. However, previous studies failed to consider using a dedicated dialogue system for the HRI system~\cite{liu2016multimodal,hong2020multimodal} or did not use different multimodal reactive emotion expressions for the robot~\cite{li2019expressing}.

\section{Methodology}
This section will first formulate the natural language processing (NLP) problem we will solve with the RL agent based on physiological rewards. Then, we will introduce our overall robot system that can personalize the behaviors of our robot for more natural HRI scenarios.

\subsection{Problem Formulation}
In the Seq2Seq problem of NLP, the goal is to train a language model (LM) than can create a mapping between the input sequence $(x_0, \dots , x_n) \in \mathcal{X}$ and output sequence $(y_0, \dots , y_n) \in \mathcal{Y}$, where $\mathcal{X}$ and $\mathcal{Y}$ are the input and output space, respectively. Given a vocabulary $\sum$, where both $\mathcal{X}$ and $\mathcal{Y} \in \sum$, and a LM can generate sequences of tokens $(y_0, \dots , y_n) \in \mathcal{Y}$ with a probability distribution using the chain rule~\cite{bengio2003neural}:
\begin{equation}
    p(y_0, \dots , y_n) = \prod_{0 \leq i < n} p(y_i|x_0, \dots, x_{n-1}).
\end{equation}

Given the current sequences of tokens, LM can generate the probability distribution of the next token. We use the probability distribution assigned by LM as the initialization of the policy, $\pi_{\theta} = p$ with learnable parameters $\theta$, and then train $\pi_{\theta}$ via RL. A reward function $r: \mathcal{X} \times \mathcal{Y} \to \mathbb{R}$ can be defined and use RL to optimize the expected reward:
\begin{equation}
    \mathbb{E} = \mathbb{E}_{x,y}[r(x,y)].
\end{equation}

We formulate the reward $r$ as a distance-based function based on Russel’s circumplex psychological model of affect~\cite{russell1980circumplex}. The positive emotions, such as \textit{happy}, and \textit{excited} have positive rewards given by the values of arousal (A) and valence (V) of these emotions. On the contrary, the negative feelings will be assigned negative reward values. Thus, the reward can be expressed as follows:
\begin{equation}
    r(x,y) = \pm\sqrt{A^2 + V^2}.
\end{equation}

The objective of our RL task is to generate more positive responses. To achieve this, we need first to train a reward model to estimate the reward based on the generated embedding vectors from the LM. Following the setting of the previous study~\cite{xie2021empathetic}, a linear classifier is added on the top of the LM to estimate the emotion and predict the level of arousal and valence values. The reward model is optimized using this loss:
\begin{equation}
\label{eqn:loss}
    \mathcal{L}_r = \mathbb{E}_{x,y}[r(x,y)].
\end{equation}

\begin{figure}[hbpt]
\centering
\includegraphics[width=1\columnwidth]{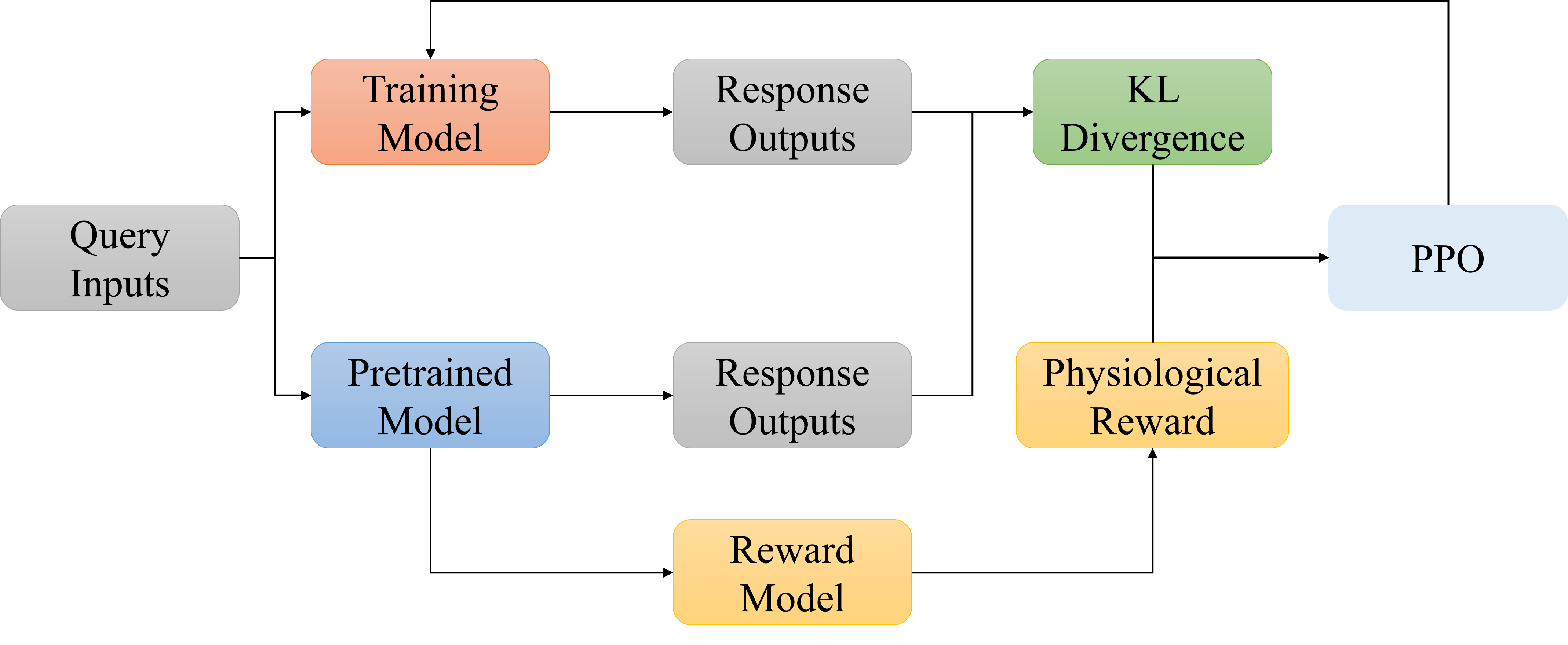} % Reduce the figure size so that it is slightly narrower than the column. Don't use precise values for figure width.This setup will avoid overfull boxes.
\caption{The workflow of the proximal policy optimization (PPO) with psychological rewards.}
\label{ppo}
\end{figure}

Following \citet{ziegler2019fine}, when fine-tuning the policy $\pi_{\theta}$ to optimize the reward via proximal policy optimization (PPO) algorithm~\cite{schulman2017proximal}, the Kullback-Leibler (KL) divergence penalty is added to prevent $\pi_{\theta}$ from changing too fast from $p$. Thus, a modified reward function is given by:
\begin{equation}
\label{eqn:reward}
    R(x,y) = r(x,y) - \beta\log\frac{\pi_{\theta}(y|x)}{p(y|x)},
\end{equation}

\noindent where $\beta$ can be chosen either fixed or adaptive using KL values~\cite{schulman2017proximal}. As can be seen in Figure~\ref{ppo}, the main objective for policy iteration will consider both psychological rewards and the KL values. The overall training process is given in Algorithm~\ref{alg:alg1}. The RL agent will optimize the LM for each sentence with annotated emotions, and by leveraging the reward function~\ref{eqn:reward}, the LM model will generate more positive responses but doesn't deviate too much compared with the original model.

\RestyleAlgo{ruled}
\begin{algorithm}
\caption{PPO with psychological reward}
\label{alg:alg1} 
Algorithm parameters: either fixed or dynamic $\beta$\;
Initialize $\pi = p$\;
Initialize $r$ to $p$ (reward model added on top of LM)\;
\ForEach{episode}{%
    % Run policy $\pi_{\theta_{\text{old}}}$ to generate sentence embeddings vectors\;
    Run language model policy $\pi$ to generate sentence embeddings vectors\;
    Optimize the reward model by the embedding vectors using loss (\ref{eqn:loss})\;
    Update the language model parameters $\theta$ via PPO with reward function (\ref{eqn:reward})\;
    % $\theta_{\text{old}}$ $\leftarrow$ $\theta$ \;
    $\theta \gets \theta'$\;
}
\KwRet{$\pi^*$}\;
\end{algorithm}

\subsection{Pre-trained Language Model}
 Generative Pre-training Transformer (GPT)~\cite{radford2018improving} is considered to be employed as the pre-trained LM in this study. GPT is transformer-based LM, where the transformer is designed purely based on the attention mechanism proposed by~\citet{vaswani2017attention}. GPT consists of multiple layers of transformer with self-attention operations, and it is also pre-trained on the large corpus. 
 
 To enable the LM the ability to recognize emotions from the dialogue, the GPT is also fine-tuned on a dialogue dataset, MELD dataset~\cite{poria2018meld}, which is targeting for the task of emotion recognition during the conversation. Each sentence from each dialogue sample of the dataset is annotated with an emotion category. The emotion labels available in the MELD dataset are \textit{Anger}, \textit{Disgust}, \textit{Sadness}, \textit{Joy}, \textit{Neutral}, \textit{Surprise}, and \textit{Fear}. The expert annotations of this dataset can provide the model the ability to assess the emotion of each user's response. The reward model, $r$ in the RL optimization process, is added as an extra layer on top of the GPT model, which is also mentioned in Algorithm~\ref{alg:alg1}.

\subsection{Human-Robot Interaction Design}
\begin{figure}[hbpt]
\centering
\includegraphics[width=1\columnwidth]{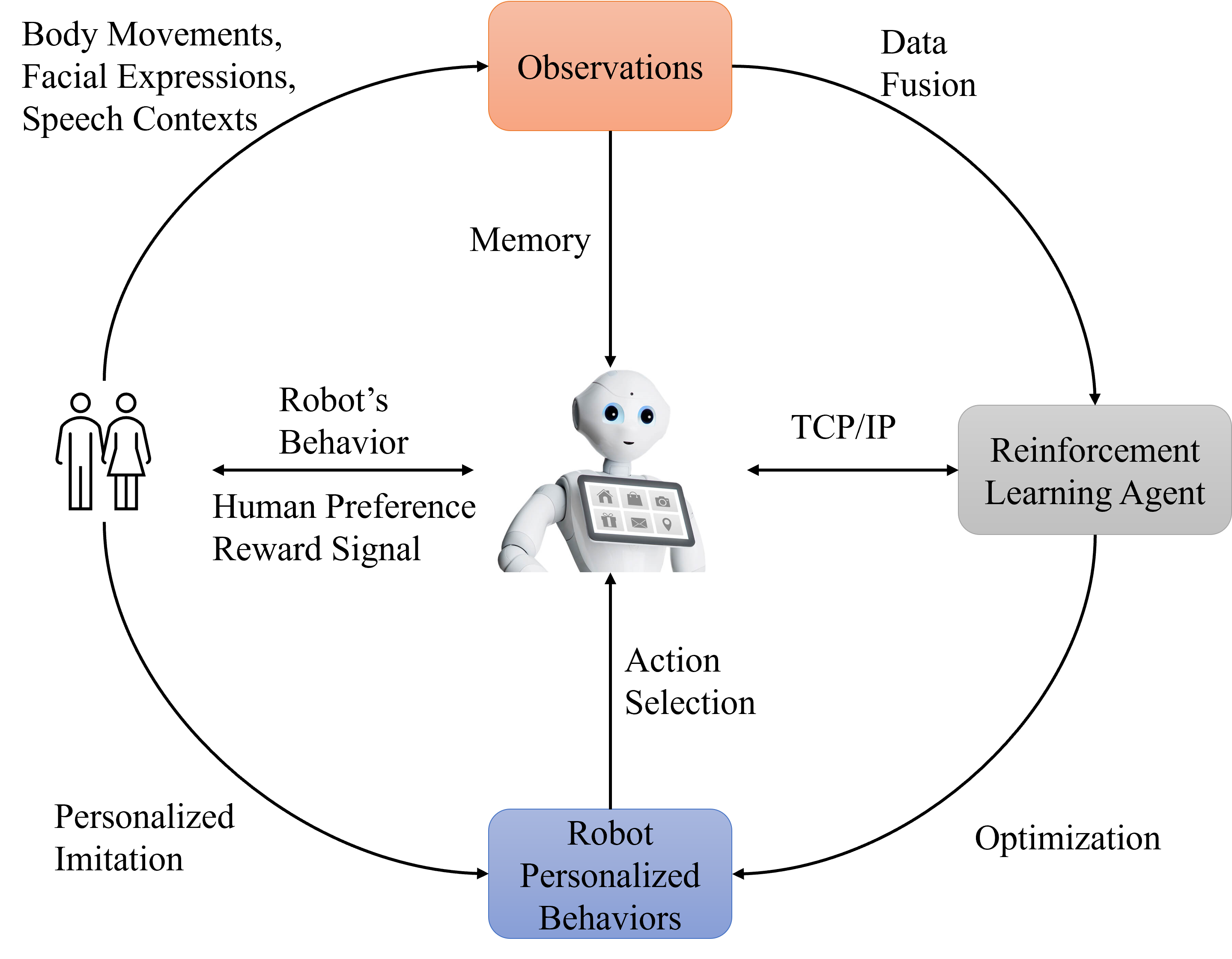} % Reduce the figure size so that it is slightly narrower than the column. Don't use precise values for figure width.This setup will avoid overfull boxes.
\caption{The proposed multimodal HRI framework with RL agent. Our robotic system (Pepper) will observe and learn human behaviors and multimodal responses, while updating behavioral policy focused on personalizing for each user.}
\label{sys}
\end{figure}

A multimodal HRI system will be designed for evaluating the RL framework. As can be seen in Figure~\ref{sys}, several modalities of observations, such as body gestures, facial expression, and speech, will be considered as the social cues and input to the system. The extracted features for the body gestures will be the human skeleton joint poses, and facial landmarks for the facial expression, and sentence embeddings for the conversational speech. The robotic platform will be the Pepper robot manufactured by the SoftBank robotics group~\cite{softbank}. Several modalities of inputs are then to be fused as part of the reward signals to the RL agent. The overall reward signal will also combine the human preferences, such as the self-evaluations from the user as the intrinsic rewards. We will adopt a self-assessment scale for the measurement of user's emotions, e.g., the self-assessment manikin~\cite{bradley1994measuring}, which is a non-verbal assessment tool that can measure affective reactions to different stimuli. The personalized robot behaviors are learned by creating the mapping between the interaction context and the affective states. The interaction context includes the textual, vocal, and gesture information from the users. The personalized robot behaviors will also be optimized through the RL reward signals.

\section{Evaluation Plan}
We plan to conduct a user study to evaluate the performance of our HRI system in generating personalized behaviors and eliciting positive emotions from the users. A two-stage study will be conducted. In the first stage, we will give the model the general knowledge of recognizing human emotions via multimodal inputs and generating feedback accordingly. To achieve this, we will use a large-scale dataset to train the model. Moreover, we will test the ability of the LM to generate positive responses after optimization by the reward signals. In the second stage, we will hold the user study to evaluate our HRI system. The agent will be personalized by utilizing the user's multimodal information and preferences. Our hypothesis is that multimodal social cues can give the robot a better ability to understand human emotional states and enhance interactive behaviors. To evaluate the user study, we propose to use Negative Attitudes towards Robots Scale (NARS)~\cite{nomura2006measurement} and Engagement~\cite{sidner2004look} questionnaires. This study hypothesizes that the proposed measures between the test group and the control group should be significantly different. A two-sided independent samples t-test would be done to compare the mean values of every extracted feature between two groups.

The experiment will be held using the Pepper humanoid robot. The user can prompt different basic topics to express their emotions during the interaction, and our empathetic dialogue agent in the system will respond accordingly. The built-in dialogue module from the Pepper robot will be used as the baseline for the control group, which only includes simple pre-defined chit-chat and rule-based contents. After the experiment, the NARS and Engagement questionnaires will be used to score both models.

\section{Conclusion}
In this study, we present a multimodal HRI framework with an RL agent. The NLP problem optimized by RL is formulated and solved by the PPO algorithm. The physiological measures will be investigated to be used as the reward signals for optimizing the RL agent. GPT is used as the pre-trained language model for the dialogue system. The proposed multimodal HRI framework will be evaluated with a two-stage user study. Different inputs modalities will be fused to recognize the user's affective states, which will also be treated as intrinsic reward signals for the RL agent. By fusing the multimodal behaviors/responses and preferences as rewards from the users, the robotic behaviors can be personalized through learning human skills and preferred feedback. We are optimistic that this work will extend the current research on social robots to provide more natural and personalized interaction capabilities, especially using multimodal HRI interaction policies and optimization through RL. For the future study, we propose using this framework to analyze and detect emotional cues from multimodal interaction data
and provide personalized intervention for adolescents with Autism Spectrum Disorder (ASD).

\section{Acknowledgment}
This research is supported by the National Science Foundation (NSF) under the grant \#1846658.

\bibliography{root}

\end{document}